\documentclass[11pt,a4paper]{article}

\usepackage[T1]{fontenc}
\usepackage[utf8]{inputenc}
\usepackage[british]{babel}
\usepackage{lmodern}
\usepackage{microtype}
\usepackage{geometry}
\usepackage{setspace}
\usepackage{booktabs}
\usepackage{array}
\usepackage{tabularx}
\usepackage{enumitem}
\usepackage{natbib}
\usepackage{hyperref}
\usepackage{xurl}
\usepackage{csquotes}
\usepackage{caption}

\setlist[itemize]{leftmargin=1.5em,itemsep=0.15em,topsep=0.25em}
\hypersetup{
  colorlinks=true,
  linkcolor=black,
  citecolor=black,
  urlcolor=black,
  pdfauthor={Simon Rudkin},
  pdftitle={From Writing Advice to Manuscript Governance: A Constrained-Transformation Framework for Academic Revision}
}

\newcommand{\AWF}{\textit{Academic Writing Filter}}

\newcommand{\doi}[1]{\href{https://doi.org/#1}{doi:#1}}

\title{From Writing Advice to Manuscript Governance:\\
A Constrained-Transformation Framework for Academic Revision}
\author{Simon Rudkin\\
\small Department of Social Statistics, University of Manchester\\
\small Manchester, United Kingdom}

\begin{document}

\maketitle

\begin{abstract}
Academic revision presents a governance problem when legitimate writing objectives conflict. Writing-process and social-science research show that revision is recursive and that wording carries epistemic and disciplinary commitments. Those literatures identify important constraints but do not settle which objective should prevail when concision, outlet form or stylistic variation conflicts with evidence and claim boundaries. This paper models revision as an admissibility problem under ordered constraints. Task-relative invariants, contextual constraints and mutable presentation choices define the admissible space; a claim-preservation contract links evidence and judgement to an authorised claim boundary. Four governance propositions specify preservation, precedence, context and sufficiency. The Academic Writing Filter implements the model by screening candidate revisions; it is not a universal definition or classifier of good academic writing. The contribution is a technology-independent mechanism for changing presentation without silently changing the research argument.
\end{abstract}


\section{Academic revision as a constrained transformation}

Writing-process research has long rejected a linear model in which ideas are completed first and prose is polished afterwards. Flower and Hayes model writing as a recursive, goal-directed process in which planning, translating and reviewing interact with the task environment and the writer's evolving goals \citep{flowerhayes1981}. Revision research likewise distinguishes local editing from changes that alter meaning, organisation and the writer's representation of the text \citep{sommers1980,faigleywitte1981,fitzgerald1987}. These accounts establish that revision can reorganise an argument rather than merely smooth sentences.

Social-science scholarship makes the problem more specific because wording can carry epistemic and disciplinary commitments. Hedges and boosters communicate degrees of commitment, caution and certainty, and that their use varies across disciplinary communities \citep{hyland1996,hyland1998}. Organisation studies further shows that contribution and theory are constructed through argument rather than supplied by citations, data or novelty claims alone \citep{lockegoldenbiddle1997,whetten1989,suttonstaw1995,alvessonsandberg2011}. A revision can therefore alter more than style when it changes a qualifier, collapses a conceptual distinction or rewrites the relationship between evidence and contribution.

Revision becomes a governance problem when individually reasonable objectives conflict. Concision may compete with a qualification needed to delimit inference. Stylistic variation may compete with a technical distinction whose repeated wording is analytically necessary. An outlet convention may shape presentation but cannot make the underlying research design support a stronger claim. Treating every instruction as an equivalent writing preference leaves those conflicts unresolved.

This paper conceptualises revision as an \emph{admissibility problem under ordered constraints}. A proposed revision is admissible only when it preserves properties fixed for the current revision task, satisfies applicable scholarly-context constraints and changes presentation within the remaining permissible space. Three classes of manuscript property structure the model: task-relative invariants, contextual constraints and mutable presentation choices. Reporting and discourse scholarship already establishes that scholarly communities differ in what must be reported, how knowledge claims are qualified and how conventions are adapted across methods and disciplines \citep{aera2006,levitt2018,lupiaelman2014,nosek2015}. Context therefore cannot be reduced to a universal style layer applied after the argument has been formed.

The term \emph{manuscript governance} is used throughout this paper in a deliberately narrow operational sense: the allocation of decision rights among competing revision constraints. Governance specifies what must be preserved, what the scholarly context constrains, what may be optimised for clarity or economy, and which requirement prevails when two legitimate revision objectives conflict. No institutional, editorial or research-governance theory is claimed.

Four linked governance propositions define the model rather than making empirical claims about its effectiveness.

\begin{enumerate}[label=\textbf{P\arabic*:},leftmargin=3.3em,itemsep=0.45em]
\item \textbf{Preservation.} When the underlying research argument is held fixed, a presentation revision is admissible only if claim-defining invariants are preserved.
\item \textbf{Precedence.} Revision constraints are heterogeneous: evidential and inferential boundaries have decision rights over lower-order stylistic or house preferences when the two conflict.
\item \textbf{Context.} Discipline, manuscript type and outlet constrain admissible expression, but those contextual requirements cannot legitimately override verified evidence or authorised inferential boundaries.
\item \textbf{Sufficiency.} Economy is a lower-order presentation objective; deletion is not an improvement when it removes context, conceptual distinction or qualification required to warrant or interpret the contribution.
\end{enumerate}

Existing literatures supply important parts of this architecture without being treated here as incomplete precursors. Cognitive models explain how writing and revision proceed \citep{flowerhayes1981}; social-science scholarship explains how contribution, theory and epistemic commitment are constructed in scholarly texts \citep{lockegoldenbiddle1997,suttonstaw1995,hyland1998}; and reporting standards specify warrants, transparency and discipline-sensitive expectations \citep{aera2006,levitt2018,lupiaelman2014,nosek2015}. A recent AI-specific framework also restricts generative AI to post-draft diagnostic auditing in order to preserve authorial judgement and epistemic control \citep{nussbaumbekerman2026}. This paper's narrower conceptual claim concerns the relation among these kinds of requirement during revision: task-relative invariants define an authorised claim boundary, contextual constraints shape admissible expression, and an explicit precedence hierarchy determines which revision objective yields when requirements conflict.

An operational implementation, the \AWF, makes that governance logic usable \citep{rudkin2026filter}. The term \emph{filter} is used operationally: candidate revisions are screened against evidence, the claim-preservation contract, contextual scholarly requirements and lower-order presentation rules. It does not denote an automatic quality classifier, a universal theory of good academic writing or a claim that the general writing advice contained in the file is novel. Conceptually, the artefact remains secondary to the governance mechanism. No claim is made here that the architecture improves writing quality, publication outcomes or reviewer evaluations.

\section{Intellectual foundations}

\subsection{Writing process and the scope of governance}

Flower and Hayes describe writing as a hierarchy of recursive processes governed by goals rather than as a fixed sequence of planning, drafting and correction \citep{flowerhayes1981}. Reviewing can interrupt other processes, and writers can revise their goals as composing proceeds. Later revision scholarship similarly distinguishes surface correction from transformations that change meaning or organisation \citep{sommers1980,faigleywitte1981,fitzgerald1987}. These models explain how revision occurs and why it can be cognitively substantive.

Manuscript governance addresses a different question. Once a revision episode holds the underlying evidence and research argument fixed, which proposed changes remain admissible? Accordingly, manuscript governance does not compete with cognitive models of composing. Its starting point is a defined presentation-revision task in which some properties are temporarily fixed, others are context-dependent and the remainder stay open to optimisation. Crossing that boundary is permitted, but it should be recognised as substantive reconsideration rather than hidden inside prose editing.

\subsection{Contribution and theory in the social sciences}

Social-science scholarship cautions against treating contribution as a property that can be generated by editorial form alone. Locke and Golden-Biddle show empirically how organisation scholars construct opportunities for contribution through intertextual positioning and problematization \citep{lockegoldenbiddle1997}. Whetten distinguishes the components and boundary conditions of theoretical contribution, while Sutton and Staw argue that references, data, variables and hypotheses cannot substitute for explanatory logic \citep{whetten1989,suttonstaw1995}. Alvesson and Sandberg further distinguish conventional gap-spotting from problematization that challenges assumptions in an existing literature \citep{alvessonsandberg2011}.

These arguments place a limit on what generic writing guidance can legitimately do. An operational revision filter can make a contribution visible, test whether its wording remains coherent and remove material that obscures it, but it cannot decide by stylistic rule what the substantive contribution should be. Project-specific substantive judgement must establish the authorised argument first. Manuscript governance then protects that argument during presentation revision rather than manufacturing novelty through rhetoric.

\subsection{Epistemic qualification and disciplinary discourse}

Claim strength is partly encoded in apparently small linguistic choices. Hyland's study of hedging in research articles treats tentativeness and qualification as resources for expressing uncertainty, precision and the writer's epistemic commitment \citep{hyland1996}. His cross-disciplinary analysis of hedges and boosters shows that statements of doubt and certainty also vary with the epistemological and interactional conventions of disciplinary communities \citep{hyland1998}. Qualifiers such as ``may'', ``suggests'' or ``is associated with'' can therefore be claim-defining rather than ornamental.

Disciplinary context also shapes which forms of argument and evidence are legible. Broader work on academic discourse reaches the same conclusion: scholarly language is not a neutral vocabulary layered onto otherwise identical arguments \citep{swales1990,hyland2004}. Together, these studies provide the basis for distinguishing a fixed inferential boundary from a contextual rhetorical convention. Governance must preserve the former while allowing the latter to shape expression where compatible with the evidence.

\subsection{Reporting standards, warrants and methodological pluralism}

Social-science reporting standards supply a further foundation. The American Educational Research Association (AERA) identifies \emph{sufficiency of warrants} and \emph{transparency} as overarching principles, while explicitly rejecting literal checklist compliance and allowing different methodological traditions to vary in form and order of presentation \citep{aera2006}. Levitt and colleagues develop qualitative and mixed-method reporting standards that are sensitive to diverse inquiry traditions and intended for use by authors, reviewers and editors across psychology and potentially the wider social sciences \citep{levitt2018}. Levitt and colleagues also document a practical source of conflict: qualitative authors can face manuscript expectations or rhetorical forms that are inappropriate for their methods.

Political-science and open-science frameworks make a similar distinction between shared principles and contextual implementation. Lupia and Elman describe data access and research transparency as general commitments that different research communities instantiate through their own standards of warrant and legitimacy \citep{lupiaelman2014}. TOP uses modular standards and explicitly allows different levels of adoption because not every standard is equally applicable across disciplines and journals \citep{nosek2015}. Together, these frameworks specify important reporting and transparency constraints. Neither is designed to allocate decision rights among a project's authorised claim boundary, outlet demands, disciplinary rhetoric and lower-order style preferences; that cross-level relation is the narrower problem addressed here.

\subsection{Argument, evidential warrant and economy}

Relations between propositions and warrants provide the immediate argumentative foundation. Toulmin makes explicit the connection between claims, grounds and warrants, while \textit{The Craft of Research} translates similar logic into practical research writing \citep{toulmin2003,booth2016}. Sutton and Staw's distinction between evidence and theory reinforces the same point within the social sciences: citations or findings do not substitute for the reasoning that connects them to a claim \citep{suttonstaw1995}. A simple audit question operationalises that relation within the \AWF: \emph{what exact evidence licences this wording?}

Four forms of warrant are distinguished. Table~\ref{tab:warrants} makes their roles explicit. Prior scholarship can warrant inherited concepts, empirical regularities and methodological properties. Evidence produced by the current manuscript can warrant new empirical, qualitative, formal or simulation-based propositions. Authoritative primary material can warrant external factual propositions for which official statistics, legal texts, administrative documentation, archival records or technical standards are more appropriate than journal articles. Transparent analytic warrant can support a definition, a logical implication or a design principle introduced and reasoned within the paper itself. Evidence is therefore not synonymous with citation. Responsible referencing requires the cited source to support the proposition for which it is cited \citep{penders2018}.

\begin{table}[htbp]
\centering
\small
\caption{Four warrant types used by the Academic Writing Filter}
\label{tab:warrants}
\begin{tabularx}{\textwidth}{@{}>{\raggedright\arraybackslash}p{0.23\textwidth}>{\raggedright\arraybackslash}X>{\raggedright\arraybackslash}p{0.22\textwidth}@{}}
\toprule
Warrant type & Appropriate use & Typical citation status \\
\midrule
Academic literature & Prior concepts, arguments, findings and methodological claims & Cite the scholarly source that supports the proposition. \\
Manuscript-generated evidence & New empirical, qualitative, formal or simulation-based propositions established in the current paper & Point to the relevant result, analysis, derivation or evidence; external citation is not a substitute. \\
Authoritative primary evidence & External facts best supported by legislation, official statistics, administrative or archival records, policies or technical standards & Cite the authoritative primary source. \\
Transparent analytic warrant & Definitions, logical implications and design principles introduced and reasoned explicitly in the paper & Make ownership and reasoning explicit; do not add decorative citations merely for density. \\
\bottomrule
\end{tabularx}
\end{table}

Cochrane's writing advice supplies an operational rather than epistemic layer. Cochrane advocates a concrete contribution, newspaper-style organisation, rapid movement to the intellectual payoff, economy, active prose and removal of research-process narrative \citep{cochrane2005}. Editorial and style guidance likewise favours direct, reader-centred prose \citep{naturemethods2017,sword2012}. Such advice is valuable inside the admissible revision space, but it cannot determine whether a design identifies a causal effect, whether a qualitative interpretation is warranted or whether a conceptual distinction may be removed. Those judgments arise from the research and the disciplinary context.

\subsection{The closest AI-specific neighbour}

Nussbaum and Bekerman provide the closest framework-level neighbour in the assembled literature \citep{nussbaumbekerman2026}. Nussbaum and Bekerman require a human-authored draft before generative AI enters the manuscript-revision process and restrict AI to diagnostic post-draft auditing. Authors retain epistemic authority while AI probes coherence, alignment and reconstructability; interpretation and evidentiary significance remain non-delegable scholarly judgments. At the level of human--AI interaction, that framework therefore addresses a closely related preservation problem.

The manuscript-governance framework shares the concern with epistemic authority but has a different unit of analysis. Manuscript governance is technology-independent: a proposed revision may originate with the author, a collaborator, a reviewer, an editor or an AI system. Governance instead asks whether that revision respects the authorised claim boundary and the precedence hierarchy. No human-first drafting requirement is built into the conceptual model. Nussbaum and Bekerman's post-draft diagnostic constraint can instead be understood as a specialised policy for AI participation, while manuscript governance specifies the more general constraints against which any accepted revision is judged.

\section{A conceptual model of manuscript governance}

\subsection{Task-relative invariants, contextual constraints and mutable choices}

Three manuscript-property classes organise revision. An \emph{invariant} is a property treated as fixed for a specified presentation-revision episode in which the underlying evidence and research argument are held constant; the term does not imply that a sentence or interpretation can never change. Reconsidering an invariant is possible, but doing so marks a move from presentation revision into substantive reconsideration. Table~\ref{tab:classes} defines the three classes and the different revision principles attached to them.

\begin{table}[htbp]
\centering
\caption{Three classes of manuscript property}
\label{tab:classes}
\begin{tabularx}{\textwidth}{@{}>{\raggedright\arraybackslash}p{0.22\textwidth}>{\raggedright\arraybackslash}X>{\raggedright\arraybackslash}X@{}}
\toprule
Class & Examples & Revision principle \\
\midrule
Task-relative invariants & Verified facts; mathematical/statistical correctness; inferential status; required qualifications; fixed conceptual distinctions; authorised and prohibited claims & Preserve within the current presentation-revision episode; changing them requires substantive reconsideration of the research or argument. \\
Contextual constraints & Discipline; manuscript type; accepted field terminology; outlet requirements; genre conventions; editor/referee instructions & Adapt expression to the scholarly context without overriding the task-relative invariants. \\
Mutable presentation choices & Sentence structure; paragraph order; transitions; wording; degree of signposting; house-style preferences & Optimise for clarity and economy within the admissible space left by the two higher classes. \\
\bottomrule
\end{tabularx}
\end{table}

The three classes are distinct rather than nested sets. Task-relative invariants determine what must remain fixed within the revision episode; contextual constraints govern how that fixed argument is expressed in a particular scholarly setting; mutable presentation choices remain open to optimisation only where they are compatible with the first two classes. The operational filter maps these classes onto specific controls: verified evidence and the claim-preservation contract protect invariants; binding requirements and outlet overlays implement contextual constraints; cross-project defaults and local house-style preferences govern mutable choices.

Separating presentation revision from substantive reconsideration gives the architecture its operational boundary. Hyland's analyses show why this distinction matters at sentence level: changing a hedge can change the degree of commitment attached to a proposition rather than merely change its style \citep{hyland1996,hyland1998}. A writer may change an invariant after new analysis, theoretical reconsideration or a deliberate change in the paper's argument, but the change should then be recognised as substantive rather than smuggled in through prose optimisation. That distinction gives the preservation proposition an operational boundary.

\subsection{Priority when constraints conflict}

Explicit ordering resolves conflicts that cannot be settled by treating every instruction as an equivalent preference. Reporting frameworks already recognise that standards vary in applicability and strength across methods, disciplines and journals \citep{aera2006,levitt2018,nosek2015}. Manuscript governance adds a cross-level ordering. Verified evidence and mathematical or statistical correctness come first. A project-specific claim-preservation contract occupies the second priority because it records the authorised claim boundary for the current revision episode. Binding outlet, editor or referee requirements come next where they are compatible with those first two levels. An outlet-specific corpus or style overlay, cross-project general writing heuristics and local house-style preferences then govern progressively more mutable presentation choices.

\begin{table}[htbp]
\centering
\small
\caption{Priority hierarchy implemented by the Academic Writing Filter}
\label{tab:hierarchy}
\begin{tabularx}{\textwidth}{@{}>{\raggedright\arraybackslash}p{0.08\textwidth}>{\raggedright\arraybackslash}X>{\raggedright\arraybackslash}X@{}}
\toprule
Priority & Control & Function \\
\midrule
1 & Verified evidence and mathematical/statistical correctness & Determines what the manuscript can truthfully claim. \\
2 & Project claim-preservation contract & Protects authorised claims, inferential boundaries, required qualifications and frozen conceptual/terminological distinctions. \\
3 & Binding outlet, editor or referee requirements & Controls formal compliance where compatible with priorities 1--2. \\
4 & Outlet-specific corpus or style overlay & Adapts structure and rhetoric to a scholarly venue. \\
5 & Cross-project general writing heuristics/default revision rules & Supplies portable writing defaults where higher-priority constraints permit. \\
6 & Local house-style preferences & Governs local prose when higher-priority constraints permit. \\
\bottomrule
\end{tabularx}
\end{table}

The \AWF\ implements the hierarchy as a whole; only the cross-project default writing heuristics occupy Priority 5. Distinguishing the artefact from that one layer prevents inherited writing advice from being mistaken for the framework's conceptual contribution.

Decision rights, rather than preference ranking alone, structure the hierarchy. A lower-priority objective remains legitimate until it conflicts with a higher-priority constraint. Concision may therefore govern between two equally accurate formulations, but it loses decision rights when shortening deletes a required qualification. Stylistic variation may improve prose, but it loses decision rights when variation collapses a fixed technical distinction. P2 therefore turns such conflicts into explicit revision decisions rather than matters of taste.

\subsection{The claim-preservation contract}

A claim-preservation contract bridges substantive evidence and admissible wording. For a specified revision episode it records authorised claims, prohibited claims, inferential boundaries, required qualifications and frozen conceptual or terminological distinctions. Verified facts and mathematical/statistical correctness remain separately protected at the first level of the hierarchy rather than being duplicated inside the contract. Hyland's work on claim qualification provides a concrete reason to record such boundaries: apparently small changes in hedging can alter epistemic commitment \citep{hyland1996,hyland1998}. In compact form, the mechanism is: \emph{research evidence and substantive judgement $\rightarrow$ authorised claim boundary $\rightarrow$ revision decision}. Table~\ref{tab:claimcontract} gives a synthetic example.

\begin{table}[htbp]
\centering
\caption{Synthetic claim-preservation contract for an observational study}
\label{tab:claimcontract}
\begin{tabularx}{\textwidth}{@{}>{\raggedright\arraybackslash}p{0.23\textwidth}>{\raggedright\arraybackslash}X@{}}
\toprule
Element & Example \\
\midrule
Authorised claim & Programme exposure is associated with lower subsequent non-participation. \\
Prohibited claim & Programme exposure reduces subsequent non-participation. \\
Inferential boundary & The observational design supports an associational claim but does not identify a programme effect. \\
Required qualification & Exposure was not randomly assigned; the analysis therefore does not establish causal identification. \\
Frozen conceptual/terminological distinction & ``Exposure'' refers to recorded programme contact; ``participation'' refers to the later observed outcome. The terms are not interchangeable. \\
\bottomrule
\end{tabularx}
\end{table}

A proposed rewrite is tested against that record before lower-order style preferences are considered. Changing ``is associated with'' to ``reduces'' crosses from the authorised associational claim to a prohibited causal claim. Alternating ``exposure'' and ``participation'' merely to avoid repetition collapses a declared distinction. Either change can become admissible only if the underlying evidence or argument is substantively reconsidered and the contract is deliberately updated. Accordingly, the contract governs a revision episode rather than declaring that a manuscript can never change.

\subsection{Conceptual model and operational filter}

The \AWF\ is not solely a style guide because its implementation also screens proposed revisions against evidential, inferential and terminology constraints; nor is it a reporting guideline. AERA, JARS--Qual, TOP and health-research guidance specify information or standards needed for transparent reporting within particular research contexts \citep{aera2006,levitt2018,nosek2015,moher2010}; the \AWF\ does not reproduce those domain-specific items or their development processes. Outlet-agnostic core rules also distinguish it from a journal rubric. AI systems can use the file as contextual instructions, but the file is not itself an AI prompt or a substitute for substantive expertise: correctness of the underlying analysis, interpretation or proof remains outside its authority.

Conceptually, the manuscript-governance framework specifies admissible revision. Its intellectual contribution is an ordering principle for revision under conflict and a mechanism for preserving claim boundaries while presentation changes. Operationally, the public \AWF\ screens candidate changes against that specification. Its warrant taxonomy, rule table and software-neutral file format support the mechanism rather than constitute separate theoretical claims.

\section{Operationalising manuscript governance}

\subsection{First application}

First application begins with five decisions \citep{rudkin2026filter}. Users establish the broad discipline, classify the manuscript type, identify a target outlet when one exists, retrieve or create the project claim-preservation contract, and write the central contribution in one concrete paragraph. These decisions turn otherwise generic writing advice into a manuscript-specific revision environment.

Revision then proceeds across four scales \citep{rudkin2026filter}. Manuscript-level rules govern contribution and architecture. Section-level rules ask what necessary question each section answers. Paragraph-level rules require an appropriately warranted proposition or a genuine procedural/reproducibility function. Sentence-level rules favour concrete subjects, early verbs and wording that does not exceed the evidence. A final audit checks the interaction among those levels rather than assuming that locally polished sentences produce a coherent paper.

\subsection{Thirteen operational rules}

The October 2026 implementation contains thirteen default revision rules \citep{rudkin2026filter}. They are not offered as thirteen novel propositions about academic writing; their role is to operationalise the governance hierarchy by combining inherited guidance, established scholarly principles and practice-derived defaults. Reproducing their full text here would duplicate the public artefact; Table~\ref{tab:rules} identifies their function and epistemic status.

\begin{table}[htbp]
\centering
\caption{Core rules and their justificatory bases in the Academic Writing Filter}
\label{tab:rules}
\small
\begin{tabularx}{\textwidth}{@{}>{\raggedright\arraybackslash}p{0.055\textwidth}>{\raggedright\arraybackslash}p{0.27\textwidth}>{\raggedright\arraybackslash}p{0.22\textwidth}>{\raggedright\arraybackslash}X@{}}
\toprule
Rule & Short label & Justificatory basis & Principal function \\
\midrule
1 & One central contribution & Cochrane-derived / practice-derived filter rule & Keeps one headline contribution while allowing secondary insights that explain it or establish scope. \\
2 & Evidential warrant & Scholarly integrity & Requires support for contestable substantive propositions. \\
3 & Warranted paragraph proposition & Practice-derived filter rule / integrity & Gives each paragraph one necessary warranted job without treating paragraph reduction as an objective. \\
4 & Triangular organisation & Cochrane-derived / practice-derived filter rule & Places the payoff early while permitting literature-led problem framing where the field requires it. \\
5 & Disciplinary terminology & Scholarly / disciplinary & Keeps concepts and evidential verbs legible within the target field. \\
6 & Research argument, not project history & Cochrane-derived / practice-derived filter rule & Removes workflow logs and internal project shorthand from outward-facing prose. \\
7 & Journal Name Rule & Practice-derived filter rule & Prevents naming the target outlet as a substitute for demonstrating fit. \\
8 & Visible work at section and sentence level & Cochrane-derived / practice-derived filter rule & Favours necessary sections, concrete subjects, early verbs and direct prose. \\
9 & Literature positions the argument & Cochrane-derived / scholarly & Uses literature argumentatively and rechecks coverage after substantive deletion. \\
10 & Minimum necessary method/theory & Cochrane-derived / practice-derived filter rule & Compresses repetition without removing conceptual distinctions required by the contribution. \\
11 & Evidence-to-claim calibration & Scholarly integrity & Distinguishes claim classes and requires a warrant appropriate to each rather than treating them as a universal ladder. \\
12 & Robustness answers named threats & Scholarly / practice-derived filter rule & Treats sensitivity analysis as a response to specific objections rather than a catalogue. \\
13 & Limitations as scope boundaries & Scholarly / practice-derived filter rule & Defines constrained claims and what evidence could cross a boundary without requiring a future-work programme for every limitation. \\
\bottomrule
\end{tabularx}
\end{table}

Table~\ref{tab:rules} reports justificatory bases rather than mutually exclusive classes: a rule may draw on Cochrane-derived guidance, broader scholarly principles and practice-derived filter rules at the same time. Distinguishing those bases prevents a local preference from masquerading as a universal epistemic rule. A heuristic against habitual openings with ``The'', ``This'', ``It'', ``There'', ``They'' or ``Their'', for example, encourages concrete subjects. Readability or disciplinary convention can override that heuristic when either provides a better reason. A prohibition on converting a descriptive association into a causal effect does not have the same optional status.

Several rules deliberately inherit rather than originate familiar social-science principles. Contribution framing is already a substantive problem in organisation studies \citep{lockegoldenbiddle1997,whetten1989,alvessonsandberg2011}; references and data cannot stand in for theoretical logic \citep{suttonstaw1995}; and reporting standards already require warrants, transparency and method-sensitive presentation \citep{aera2006,levitt2018}. Within the present manuscript-governance framework, the \AWF\ places those requirements inside one revision hierarchy rather than relabelling established disciplinary knowledge as new writing guidance.

Post-revision sufficiency follows from the hierarchy rather than operating as an exception to it. Economy and paragraph reduction belong to mutable presentation choices, whereas a required qualification can be invariant and disciplinary positioning can be a contextual constraint. A substantial deletion therefore requires a check that a lower-order optimisation has not damaged a higher-order requirement. After major compression, the October 2026 implementation checks literature coverage, necessary conceptual distinctions, Discussion depth and the navigation value of exhibits \citep{rudkin2026filter}. Literature-led problem framing, short warranted paragraphs and navigation exhibits can survive when they perform a higher-priority argumentative or contextual function. Under the sufficiency proposition, shorter is better only inside the admissible revision space.

\subsection{Implementation as a portable artefact}

Markdown provides an inspectable, portable specification independent of a particular model or word processor. Users can compare dated releases, add discipline-specific terminology or an outlet overlay, and supply the same rules to a human collaborator or an AI-assisted system. Its role remains implementational: plain text exposes governance rules to inspection and adaptation but does not create the conceptual contribution.

\section{Worked transformations}

Synthetic examples show how ordered constraints govern revision decisions; they make no empirical claims about the substantive topics mentioned.

\subsection{Claim calibration}

Consider a hypothetical observational study whose draft states:

\begin{quote}
Exposure to the programme reduced subsequent non-participation.
\end{quote}

Under the claim-preservation contract in Table~\ref{tab:claimcontract}, a causal interpretation is prohibited. An admissible revision therefore reads:

\begin{quote}
Programme exposure was associated with lower subsequent non-participation.
\end{quote}

Adding ``may'', ``suggests'' or ``is consistent with'' would not by itself repair an unsupported causal claim. Hyland shows that hedges legitimately calibrate epistemic commitment, but they do not supply evidence that the design lacks \citep{hyland1996,hyland1998}. Causal wording should therefore be aligned with the causal question and the design used to identify it rather than softened through euphemism \citep{hernan2018}. Higher-order evidential constraints determine the claim class before sentence style is optimised.

\subsection{Conflict resolution: concision versus qualification}

Suppose the authorised claim is followed by a qualification required by the contract:

\begin{quote}
Programme exposure was associated with lower subsequent non-participation. Because exposure was not randomly assigned, the association does not identify a programme effect.
\end{quote}

A compression pass proposes deleting the second sentence because the first is already associational. Concision is a legitimate objective, but the contract marks the qualification as required for interpretation. Because the qualification is an invariant, the second sentence remains. Governance does not declare concise prose undesirable; it specifies that concision has decision rights only among formulations that preserve the authorised claim boundary.

\subsection{Conflict resolution: variation versus a fixed distinction}

A copy-editing pass may also propose alternating ``exposure'' and ``participation'' to avoid repetition. Table~\ref{tab:claimcontract} defines those terms as different analytical objects. Lexical variation is therefore rejected even when it would improve surface rhythm. Repetition remains because a house preference for variation cannot override a claim-defining distinction. Rejecting the substitution illustrates why an unranked checklist is insufficient when two reasonable writing objectives point in different directions.

\subsection{From confessional limitation to scope boundary}

A generic limitations paragraph might say:

\begin{quote}
The study does not identify individual-level mechanisms. Future research should examine these mechanisms.
\end{quote}

A governed revision instead links the limitation to the authorised claim:

\begin{quote}
The area-level design identifies geographic patterning rather than individual behavioural pathways. That scope is sufficient for the paper's claim about spatial concentration, but it does not identify the mechanisms that produce individual participation decisions. Linked individual-level data could test those pathways directly.
\end{quote}

Work on limitations reporting likewise argues that limitations should explain consequences for interpretation rather than function as ritualised disclosures \citep{tsagkaris2026}. Governance adds a further condition: future work cannot rescue a present validity problem. If missing evidence is necessary to establish the central claim, the design or the claim must change in the current manuscript; otherwise, the limitation defines the boundary of what the present design authorises.

\section{Human and AI-assisted use}

Technology independence allows the same hierarchy to govern several revision settings. Individual researchers, supervisors, co-authors and writing groups can use the hierarchy to test whether a proposed rewrite changes the authorised claim, conflicts with disciplinary terminology or applies a lower-priority preference too aggressively. An AI-assisted writing system can use the same specification as contextual instructions, but the system does not acquire authority to redefine the evidence or claim-preservation contract.

Nussbaum and Bekerman offer a complementary, AI-specific design \citep{nussbaumbekerman2026}. Their ``conversational companion'' begins from a human-authored draft and limits generative AI to post-draft diagnostic work on coherence, alignment and reconstructability. A further distinction between delegable assistance and non-delegable interpretation reserves decisions about evidentiary meaning and conclusions to human scholarly judgement. Manuscript governance accepts the same general concern with epistemic responsibility but does not prescribe a particular agent or drafting sequence. Its unit of analysis is the revision decision: regardless of who proposes a change, acceptance depends on the evidence, the claim-preservation contract and the contextual hierarchy.

AI-assisted scientific writing can support organisation, drafting and language editing while creating risks of inaccuracy, inappropriate attribution and over-reliance on fluent output \citep{salvagno2023}. Current ICMJE guidance places responsibility for AI-assisted content on human authors and requires disclosure of AI-assisted technologies used in manuscript production \citep{icmje2026}. Applicable disclosure requirements remain unchanged by use of the \AWF. Its narrower role is to make revision constraints explicit and repeatable across editing contexts; no text file guarantees compliance by a human or a model.

\section{Boundaries and open release}

As a design artefact, the \AWF\ operationalises the manuscript-governance model but is not a validated reporting guideline or a universal standard of academic writing quality. No effectiveness study or consensus process has been undertaken. No present evidence establishes that using the filter improves manuscript quality, publication probability, reviewer agreement or writing speed. Those outcomes could be evaluated separately, but they are not part of the present claim.

No claim of disciplinary universality is made. Social-science reporting and discourse research shows why such a claim would be inappropriate: standards, rhetorical practices and evidential conventions differ across methodological and disciplinary communities \citep{hyland1998,aera2006,levitt2018,lupiaelman2014,nosek2015}. Discipline, manuscript type and outlet therefore enter as contextual constraints rather than after-the-fact style adjustments. Contextual constraints must shape expression while preserving higher-order evidence and claim boundaries.

Nor does this paper claim to invent revision, contribution framing, hedging, transparent reporting or AI-assisted auditing. Those elements have established literatures \citep{flowerhayes1981,lockegoldenbiddle1997,hyland1996,aera2006,nussbaumbekerman2026}. This paper claims a narrower contribution: an explicit integration in which a task-relative claim boundary and a precedence hierarchy govern what may change when legitimate revision requirements conflict.

Cochrane provides the principal architectural starting point for the operational writing rules \citep{cochrane2005}, but the manuscript-governance framework remains independent. Controls over evidential warrant, discipline, claim preservation, terminology and limitations combine broader scholarly principles with practice-derived rules. No endorsement by John H. Cochrane is implied.

Open release is intended to make the implementation inspectable and contestable. A stable repository or DOI can preserve the dated Markdown file and this working paper after the provenance and licence review is complete. Release mechanics are deliberately secondary to the conceptual claim: manuscript governance concerns ordered revision constraints regardless of where the implementation is stored.

\section{Conclusion}

Revision research, social-science writing scholarship and reporting standards establish that rewriting is recursive, epistemically consequential and context-dependent \citep{flowerhayes1981,hyland1998,aera2006,levitt2018}. Manuscript governance addresses the narrower problem created when those legitimate requirements conflict during a presentation-revision episode. Prose improvement alone cannot decide whether concision should remove a qualification, whether stylistic variation may alter a fixed term, or whether an outlet convention may strengthen a claim beyond the evidence.

The model created in this paper separates task-relative invariants from contextual constraints and mutable presentation choices, then assigns decision rights among them. A claim-preservation contract operationalises the central mechanism: evidence and substantive judgement define an authorised claim boundary; proposed wording is tested against that boundary before outlet form, economy or local house style are optimised. Changing an invariant remains possible, but doing so marks substantive reconsideration rather than ordinary presentation revision. That ordering also explains the post-revision sufficiency audit because lower-order compression cannot count as improvement when it removes a higher-order qualification, conceptual distinction or disciplinary requirement.

A portable Markdown file, the \AWF, implements this architecture by screening candidate revisions through the hierarchy. Its thirteen rules combine inherited writing guidance, established scholarly principles and practice-derived controls; they are not presented as novel general writing guidance and are not the paper's principal intellectual claim. Conceptually, the contribution is the ordering mechanism that connects claim preservation, context and presentation choice under conflict. That mechanism is technology-independent, although AI-specific frameworks such as Nussbaum and Bekerman's post-draft diagnostic model illustrate one setting in which explicit boundaries around epistemic responsibility are especially salient \citep{nussbaumbekerman2026}. Open release can make the mechanism available for use, criticism and adaptation without claiming an effectiveness that has not been demonstrated.

\section*{Availability}

The October 2026 release of the companion implementation, \nolinkurl{Academic_Writing_Filter_10_2026.md}, is available from the public project page at \url{https://sites.google.com/view/simonrudkin/ai}.

\section*{AI-assisted development statement}

OpenAI's ChatGPT was used during development of the \AWF\ to consolidate successive drafts, identify candidate literature, test internal consistency and assist with preparation and revision of this working paper and its \LaTeX\ source. Candidate sources used in the present revision were retained only after full-text copies were obtained and inspected. Author review and revision followed, and responsibility for the arguments, citations and final text remains with the author.

\section*{Acknowledgement}

Cochrane's \textit{Writing Tips for Ph.D. Students} provides the operational writing rules' principal starting point. Cochrane did not participate in the development of the \AWF\ and no endorsement is implied.


\begin{thebibliography}{99}

\bibitem[Cochrane(2005)]{cochrane2005}
Cochrane, J. H. (2005).
\newblock \textit{Writing Tips for Ph.D. Students}.
\newblock Graduate School of Business, University of Chicago.
\newblock \url{https://www.johnhcochrane.com/research-all/writing-tips-for-phd-studentsnbsp}.

\bibitem[Toulmin(2003)]{toulmin2003}
Toulmin, S. E. (2003).
\newblock \textit{The Uses of Argument}, updated ed.
\newblock Cambridge University Press, Cambridge.
\newblock \doi{10.1017/CBO9780511840005}.

\bibitem[Booth et~al.(2016)Booth, Colomb, Williams, Bizup, and FitzGerald]{booth2016}
Booth, W. C., Colomb, G. G., Williams, J. M., Bizup, J., and FitzGerald, W. T. (2016).
\newblock \textit{The Craft of Research}, 4th ed.
\newblock University of Chicago Press, Chicago.

\bibitem[Swales(1990)]{swales1990}
Swales, J. M. (1990).
\newblock \textit{Genre Analysis: English in Academic and Research Settings}.
\newblock Cambridge University Press, Cambridge.

\bibitem[Hyland(2004)]{hyland2004}
Hyland, K. (2004).
\newblock \textit{Disciplinary Discourses: Social Interactions in Academic Writing}.
\newblock University of Michigan Press, Ann Arbor.
\newblock \doi{10.3998/mpub.6719}.

\bibitem[Sword(2012)]{sword2012}
Sword, H. (2012).
\newblock \textit{Stylish Academic Writing}.
\newblock Harvard University Press, Cambridge, MA.

\bibitem[Penders(2018)]{penders2018}
Penders, B. (2018).
\newblock Ten simple rules for responsible referencing.
\newblock \textit{PLOS Computational Biology}, 14(4), e1006036.
\newblock \doi{10.1371/journal.pcbi.1006036}.

\bibitem[Nature Methods(2017)]{naturemethods2017}
Nature Methods (2017).
\newblock So you're writing a paper.
\newblock \textit{Nature Methods}, 14, 1115.
\newblock \doi{10.1038/nmeth.4532}.

\bibitem[Salvagno et~al.(2023)Salvagno, Taccone, and Gerli]{salvagno2023}
Salvagno, M., Taccone, F. S., and Gerli, A. G. (2023).
\newblock Can artificial intelligence help for scientific writing?
\newblock \textit{Critical Care}, 27, 75.
\newblock \doi{10.1186/s13054-023-04380-2}.

\bibitem[Tsagkaris and Sevik(2026)]{tsagkaris2026}
Tsagkaris, C. G. and Sevik, I. (2026).
\newblock Is there a limit to limitations? A proposal for structured limitations reporting in scientific manuscripts.
\newblock \textit{BMC Medical Research Methodology}, 26, 193.
\newblock \doi{10.1186/s12874-026-02992-0}.


\bibitem[Sommers(1980)]{sommers1980}
Sommers, N. (1980).
\newblock Revision strategies of student writers and experienced adult writers.
\newblock \textit{College Composition and Communication}, 31(4), 378--388.
\newblock \doi{10.2307/356588}.

\bibitem[Faigley and Witte(1981)]{faigleywitte1981}
Faigley, L. and Witte, S. (1981).
\newblock Analyzing revision.
\newblock \textit{College Composition and Communication}, 32(4), 400--414.
\newblock \doi{10.2307/356602}.

\bibitem[Fitzgerald(1987)]{fitzgerald1987}
Fitzgerald, J. (1987).
\newblock Research on revision in writing.
\newblock \textit{Review of Educational Research}, 57(4), 481--506.
\newblock \doi{10.3102/00346543057004481}.

\bibitem[Hernán(2018)]{hernan2018}
Hernán, M. A. (2018).
\newblock The C-word: Scientific euphemisms do not improve causal inference from observational data.
\newblock \textit{American Journal of Public Health}, 108(5), 616--619.
\newblock \doi{10.2105/AJPH.2018.304337}.

\bibitem[Moher et~al.(2010)Moher, Schulz, Simera, and Altman]{moher2010}
Moher, D., Schulz, K. F., Simera, I., and Altman, D. G. (2010).
\newblock Guidance for developers of health research reporting guidelines.
\newblock \textit{PLOS Medicine}, 7(2), e1000217.
\newblock \doi{10.1371/journal.pmed.1000217}.


\bibitem[Flower and Hayes(1981)]{flowerhayes1981}
Flower, L. and Hayes, J. R. (1981).
\newblock A cognitive process theory of writing.
\newblock \textit{College Composition and Communication}, 32(4), 365--387.

\bibitem[Hyland(1996)]{hyland1996}
Hyland, K. (1996).
\newblock Writing without conviction? Hedging in science research articles.
\newblock \textit{Applied Linguistics}, 17(4), 433--454.

\bibitem[Hyland(1998)]{hyland1998}
Hyland, K. (1998).
\newblock Boosting, hedging and the negotiation of academic knowledge.
\newblock \textit{Text}, 18(3), 349--382.

\bibitem[American Educational Research Association(2006)]{aera2006}
American Educational Research Association (2006).
\newblock Standards for reporting on empirical social science research in AERA publications.
\newblock \textit{Educational Researcher}, 35(6), 33--40.
\newblock \doi{10.3102/0013189X035006033}.

\bibitem[Levitt et~al.(2018)Levitt, Bamberg, Creswell, Frost, Josselson, and Su\'arez-Orozco]{levitt2018}
Levitt, H. M., Bamberg, M., Creswell, J. W., Frost, D. M., Josselson, R., and Su\'arez-Orozco, C. (2018).
\newblock Journal article reporting standards for qualitative primary, qualitative meta-analytic, and mixed methods research in psychology: The APA Publications and Communications Board task force report.
\newblock \textit{American Psychologist}, 73(1), 26--46.
\newblock \doi{10.1037/amp0000151}.

\bibitem[Lupia and Elman(2014)]{lupiaelman2014}
Lupia, A. and Elman, C. (2014).
\newblock Openness in political science: Data access and research transparency.
\newblock \textit{PS: Political Science \& Politics}, 47(1), 19--42.
\newblock \doi{10.1017/S1049096513001716}.

\bibitem[Nosek et~al.(2015)Nosek, Alter, Banks, Borsboom, Bowman, Breckler, Buck, Chambers, Chin, Christensen, Contestabile, Dafoe, Eich, Freese, Glennerster, Goroff, Green, Hesse, Humphreys, Ishiyama, Karlan, Kraut, Lupia, Mabry, Madon, Malhotra, Mayo-Wilson, McNutt, Miguel, Paluck, Simonsohn, Soderberg, Spellman, Turitto, VandenBos, Vazire, Wagenmakers, Wilson, and Yarkoni]{nosek2015}
Nosek, B. A. et al. (2015).
\newblock Promoting an open research culture.
\newblock \textit{Science}, 348(6242), 1422--1425.
\newblock \doi{10.1126/science.aab2374}.

\bibitem[Locke and Golden-Biddle(1997)]{lockegoldenbiddle1997}
Locke, K. and Golden-Biddle, K. (1997).
\newblock Constructing opportunities for contribution: Structuring intertextual coherence and ``problematizing'' in organizational studies.
\newblock \textit{Academy of Management Journal}, 40(5), 1023--1062.

\bibitem[Whetten(1989)]{whetten1989}
Whetten, D. A. (1989).
\newblock What constitutes a theoretical contribution?
\newblock \textit{Academy of Management Review}, 14(4), 490--495.

\bibitem[Sutton and Staw(1995)]{suttonstaw1995}
Sutton, R. I. and Staw, B. M. (1995).
\newblock What theory is not.
\newblock \textit{Administrative Science Quarterly}, 40(3), 371--384.
\newblock \doi{10.2307/2393788}.

\bibitem[Alvesson and Sandberg(2011)]{alvessonsandberg2011}
Alvesson, M. and Sandberg, J. (2011).
\newblock Generating research questions through problematization.
\newblock \textit{Academy of Management Review}, 36(2), 247--271.

\bibitem[Nussbaum and Bekerman(2026)]{nussbaumbekerman2026}
Nussbaum, M. and Bekerman, Z. (2026).
\newblock Generative AI as a conversational companion: A design-oriented post-draft framework for preserving authorial judgment and creative agency in scientific manuscripts.
\newblock \textit{Thinking Skills and Creativity}, 62, 102335.
\newblock \doi{10.1016/j.tsc.2026.102335}.

\bibitem[ICMJE(2026)]{icmje2026}
International Committee of Medical Journal Editors (2026).
\newblock Recommendations: Use of AI by authors.
\newblock \url{https://www.icmje.org/recommendations/browse/artificial-intelligence/ai-use-by-authors.html}.
\newblock Accessed 3 October 2026.

\bibitem[Rudkin(2026)]{rudkin2026filter}
Rudkin, S. (2026).
\newblock \textit{Academic Writing Filter}.
\newblock October 2026 release. \url{https://sites.google.com/view/simonrudkin/ai}.

\end{thebibliography}
\end{document}